\documentclass[11pt]{article}
\usepackage{authblk}
\usepackage{acl2016}
\usepackage{times}
\usepackage{latexsym}
\usepackage{graphicx}
\usepackage{caption}
\usepackage{multirow}
\usepackage{amssymb}

\usepackage{array}

\makeatletter
\newcommand{\@BIBLABEL}{\@emptybiblabel}
\newcommand{\@emptybiblabel}[1]{}
\makeatother
\usepackage{hyperref}

\usepackage{array}
\newcolumntype{L}{>{\centering\arraybackslash}m{5cm}}

\aclfinalcopy 

\title{Character-level and Multi-channel Convolutional Neural Networks for Large-scale Authorship Attribution}

\author[1,2]{Sebastian Ruder}
\author[2]{Parsa Ghaffari}
\author[1]{John G. Breslin}
\affil[1]{Insight Centre for Data Analytics}
\affil[ ]{National University of Ireland, Galway}
\affil[ ]{\tt \{sebastian.ruder,john.breslin\}@insight-centre.org}
\affil[2]{Aylien Ltd.}
\affil[ ]{Dublin, Ireland}
\affil[ ]{\tt \{sebastian,parsa\}@aylien.com}

\date{}

\begin{document}

\maketitle

\begin{abstract}
Convolutional neural networks (CNNs) have demonstrated superior capability for extracting information from raw signals in computer vision. Recently, character-level and multi-channel CNNs have exhibited excellent performance for sentence classification tasks.
We apply CNNs to large-scale authorship attribution, which aims to determine an unknown text's author among many candidate authors, motivated by their ability to process character-level signals and to differentiate between a large number of classes, while making fast predictions in comparison to state-of-the-art approaches.
We extensively evaluate CNN-based approaches that leverage word and character channels and compare them against state-of-the-art methods for a large range of author numbers, shedding new light on traditional approaches. We show that character-level CNNs outperform the state-of-the-art on four out of five datasets in different domains. Additionally, we present the first application of authorship attribution to reddit. Finally, we release our new reddit and Twitter datasets for further research.
\end{abstract}

\section{Introduction}

State-of-the-art methods in authorship attribution, which aims to determine an unknown text's author among a set of candidate authors, rely on low-level information such as character n-grams \cite{Frantzeskou2007}. Recent approaches \cite{Koppel2011} focus on large-scale authorship attribution for thousands of authors, but are expensive during prediction, which is a deficit in on-line scenarios for purposes of targeted marketing, copyright enforcement, writing support, and search relevance, among others \cite{Potthast}. Furthermore, besides stylistic information, word-level topical information has been shown to be relevant for authorship attribution \cite{Seroussi2011}.

Simultaneously, convolutional neural networks (CNNs) have achieved remarkable successes in computer vision \cite{Krizhevsky2012} and speech recognition \cite{Abdel-Hamid2012} and have been found particularly suitable for extracting information from just such low-level signals. They have also been shown to be effective for various NLP tasks \cite{Collobert2011a} and have achieved state-of-the-art in several sentence classification tasks \cite{Kim2014}. 
Most neural networks and CNNs in NLP perform convolutions on the word level using pre-trained word embeddings \cite{Mikolov2013d}. Recent approaches employ convolutions over characters \cite{Zhang2015d}.

We apply CNNs to the task of authorship attribution for four reasons: a) They have been shown to be excellent at leveraging character-level signals, which have been found to be indicative of authorial style \cite{Stamatatos2009}; b) they have excelled at differentiating between a large number of classes \cite{Krizhevsky2012}, which is key for large-scale authorship attribution; c) prediction is fast; and d) a combination of word and character input channels enables them to take topical information into account.

Our main contributions are the following:
\begin{itemize}
	\item We present state-of-the-art results for large-scale authorship attribution for four out of five datasets in different domains using character-level convolutional neural networks.
	\item We evaluate combinations of different CNN input channels and propose a novel model that combines character and word channels.
	\item We compare CNN approaches against traditional approaches for a large range of author numbers and discuss merits and improvements.
	\item We present the first application of authorship attribution to reddit comments and introduce two new Twitter and reddit datasets that we make available for further research. 
\end{itemize}

\section{Related work}

Our work is inspired by two neural network architectures: multi-channel CNNs and character-level CNNs.

\textbf{Multi-channel CNNs} are pervasive in domains where the input can naturally be separated into different channels, e.g. color channels in computer vision, wave lengths in speech recognition \cite{Hoshen2014}. Natural language input is typically single-channel in the form of tokens or characters. Kim \shortcite{Kim2014} observe that a static word channel is able to encode general semantic similarities, while a non-static channel can be fine-tuned to the task at hand and improves performance on some datasets.

\textbf{Character-level CNNs} have been shown to outperform traditional classification methods on large-scale datasets \cite{Zhang2015d}. Their CNNs, however, require tens of thousands of per-class examples and thousands of training epochs, while our datasets only contain a few hundred examples per class.

The use of most other character-level CNNs is motivated by the desire to leverage sub-word information (e.g. morphemes) to which word-level CNNs are blind: Kim et al. \shortcite{Kim2016} feed the output of a character-level CNN to a recurrent neural language model and improve performance particularly for morphologically rich languages.

Santos and Zadrozny \shortcite{Santos2014} use a CNN that associates a character embedding produced by a CNN for each word with its word representation to improve POS tagging performance for English and Portuguese, while Santos and Guimar\~{a}es \shortcite{Santos2015} use the same network to boost results for named entity recognition. In contrast to their approach, we do not aim to detect morphological information because inter-word features such as punctuation and white space are very relevant to authorship. Rather, we treat characters as a distinct input channel as our goal is to learn to identify discrete word and character patterns and to associate them with each other.

\textbf{Authorship attribution} is the task of identifying an unknown text's author among a set of candidate authors with applications ranging from plagiarism detection to Forensic Linguistics. The key notion behind statistical authorship attribution is that measuring textual features enables distinction between texts written by different authors \cite{Stamatatos2009}. These features range from indicators of content divergence between authors such as bag-of-words to stylometric features that reflect an author's unique writing patterns, e.g. use of punctuation marks, emoticons, whitespace, etc. \cite{Sapkota2011}, and character and word n-grams \cite{Schwartz2013}.

Deep learning research has largely neglected authorship attribution; related work has instead focused on modeling an author's style: Kiros et al. \shortcite{Kiros2014a} condition word embeddings on attributes such as style and predict an author's age, gender, and industry. Zhu et al. \shortcite{Zhu2015a} transform image captions into book sentences by subtracting the 'style'.
State-of-the-art authorship attribution algorithms have to handle possibly thousands of candidate authors and a limited number of examples per author in real-world applications but require CPU-days for prediction as they calculate pairwise distances between feature subsets \cite{Koppel2011}. Simultaneously, character n-grams have proven to be the single most successful feature \cite{Frantzeskou2007}.
Finally, Potthast et al. \shortcite{Potthast} compare traditional approaches on small datasets, while we evaluate state-of-the-art as well as CNN-based methods for thousands of authors, thereby moving a step closer to the goal of authorship attribution at web-scale.

\begin{figure*}[]
\centering
\includegraphics[width=\linewidth]{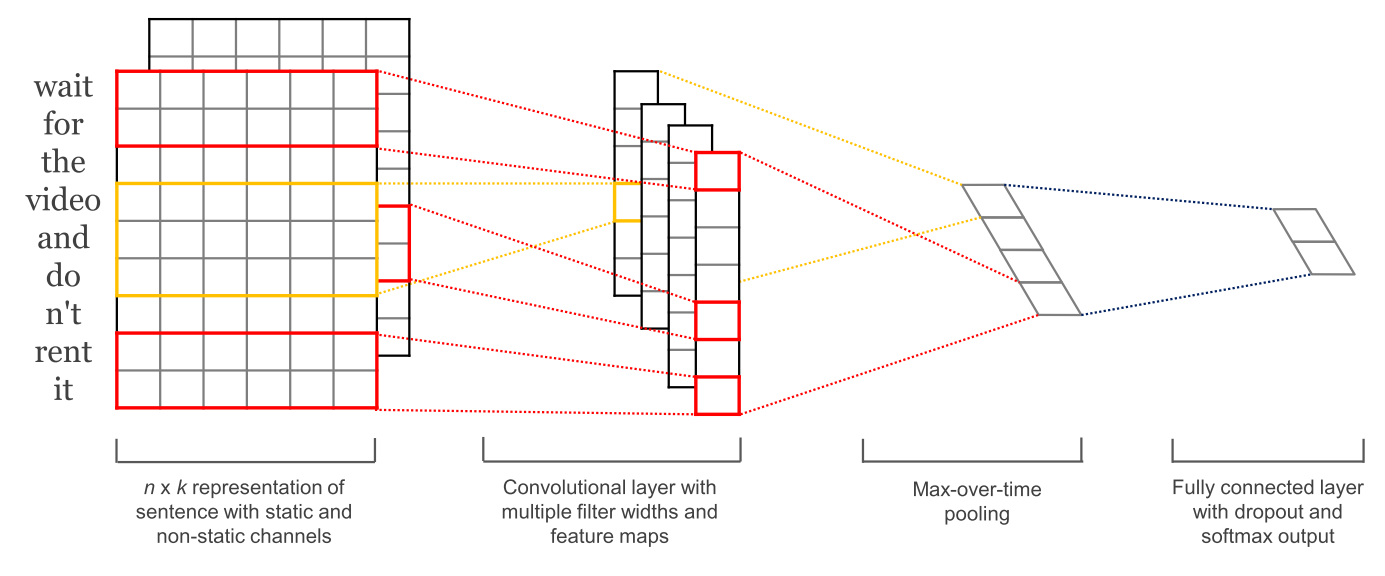}
\caption{Multi-channel CNN with two word channels from Kim et al. (2014)}
\label{fig:multi-channel-CNN}
\end{figure*}

\section{Model}

The model architecture we use is an extension of the CNN structure used by Collobert et al. \shortcite{Collobert2011a}. Its variant with two word embedding channels used by Kim \shortcite{Kim2014} is depicted in Figure \ref{fig:multi-channel-CNN}.

The model takes as input a text, which is padded to length $n$. For the character embedding channel, we represent the text as a concatenation of its character embeddings $z_{1:m}$ where $z_i \in \mathbb{R}^k$ is the $k$-dimensional vector of the $i$-th character in the text and $m$ is the character length of the text, while for the word embedding channel, we represent the text as a concatentation of its word embeddings $x_{1:n}$ where $x_i \in \mathbb{R}^k$ is the $k$-dimensional vector of the $i$-th word in the text. The convolutional layer slides filters of different window sizes over each input channel. Each filter with weights $w \in \mathbb{R}^{hk}$ generates a new feature $c_i$ for a window of $h$ characters or words according to the following operation:

\begin{equation} \label{eq:featuremap}
c_i = f(w \cdot x_{i:i+h-1} + b) 
\end{equation}

Note that $b \in \mathbb{R}$ is a bias term and $f$ is a non-linear function, ReLU \cite{Nair2010} in our case. The application of the filter over each possible window of $h$ words or characters in the sentence produces the following feature map:

\begin{equation} 
c = [c_1, c_2, ..., c_{n-h+1}]
\end{equation}

Max-over-time pooling \cite{Collobert2011a} in turn condenses this feature vector to its most important feature by taking its maximum value and naturally deals with variable input lengths.
A final softmax layer takes the concatenation of the maximum values of the feature maps produced by all filters and outputs a probability distribution over all candidate authors.

For the standard single-channel character-level CNN, we represent characters as one-hot vectors.\footnote{Using low-dimensional character embeddings as in Kim et al. \shortcite{Kim2016} decreased performance.}

In the multi-channel architecture, we apply each filter to both channels and sum the results\footnote{We have also experimented with concatenation, which produces slightly worse results.} to calculate $c_{i}$ in equation \ref{eq:featuremap}. Note that in order to combine a word and character channel, both spaces must have the same dimensionality. We thus embed characters in a space with the same number of dimensions as the word embedding space and pad the word sequence to have the same length as the character sequence.\footnote{We found that truncating the character sequence yielded worse performance.}

\section{Datasets}

We benchmark our models on the following datasets that cover a large spectrum of styles and domains:

The \textbf{Enron email dataset} contains about 0.5M emails from mostly senior managers at Enron\footnote{\url{https://www.cs.cmu.edu/~./enron/}} organized in different folders. We extract emails from the \textsc{sent} and \textsc{sent\_items} folders as these include outgoing traffic. We remove all but the email body, discarding emails that contain less than 10 tokens. This leaves us with 80,852 emails from 144 authors.

The \textbf{IMDb62 dataset} contains 62,000 movie reviews by 62 prolific users of the Internet Movie database. The dataset is made available by Seroussi et al. \shortcite{Seroussi2011} on request.

The \textbf{Blog authorship dataset} contains 681,288 blog posts of 19,320 bloggers gathered by Schler et al. \shortcite{Schler2005} from blogger.com in August 2004\footnote{\url{http://u.cs.biu.ac.il/~koppel/BlogCorpus.htm}}.

We create the \textbf{Twitter influencer dataset} in order to establish and share benchmark train and test splits for authorship attribution on a dedicated Twitter corpus\footnote{The dataset collected by Layton et al. \shortcite{Layton2010} is no longer available.}. We created the dataset by gathering a list of 4,391 celebrity and power-user influencers in 68 domains ranging from politics and tech to arts and writing and collected just over 1M tweets for these users in October and November 2015 using the AYLIEN API\footnote{\url{http://aylien.com/}}.

The \textbf{reddit gaming dataset} is a subset of the massive dataset of around 1.7 billion reddit comments from 2007 to 2015 collected via reddit's API\footnote{\url{https://www.reddit.com/dev/api}}. We control for topic by selecting comments only from the \emph{/r/Gaming} subreddit and download 2 million comments of the subreddit's most prolific users from January 2014 to May 2015 using Google BigQuery\footnote{\url{https://cloud.google.com/bigquery/}}.
We show statistics for the datasets in table \ref{tab:datasets}.

\section{Experimental setup}

\subsection{Model variations}

We test variations of our model that use different combinations of input channels together with a one-layer CNN.\footnote{We have compared against character- and word-level recurrent neural networks, but do not include them in the evaluation as their performance was not competitive.}

\begin{itemize}
	\item \textbf{CNN-word}: a CNN with a non-static word embedding channel where the vectors are modified during training using back-propagation. This is the classic word embedding CNN used e.g. by Collobert et al. \shortcite{Collobert2011a}.
	\item \textbf{CNN-char}: a CNN with a non-static character channel. A variant with  more than one convolutional layer is used by Zhang et al. \shortcite{Zhang2015d}.
	\item \textbf{CNN-word-word}: a CNN with a static and a non-static word embedding channel. This is the multi-channel CNN proposed by Kim \shortcite{Kim2014}.
	\item \textbf{CNN-word-char}: a hybrid-channel CNN with a non-static word and a non-static character channel.
	\item \textbf{CNN-word-word-char}: a hybrid-channel CNN with a static word, a non-static word and a non-static character channel.

\end{itemize}

\begin{table}[]
\centering
\begin{tabular}{l|c|c|c|c|c}
\textbf{Data} & $c$ & $D$ & $\mu$ & $\sigma$ & $d$ \\\hline
Emails & 144 & 80,852 & 561 & 889 & 175 \\
IMDb & 62 & 62,000 & 1,000 & 0 & 347 \\
Blogs & 19,320 & 681,288 & 35 & 105 & 440 \\
Twitter & 4,391 & 1,004,399 & 229 & 730 & 19 \\
reddit & 9,266 & 2,000,000 & 216 & 223 & 31
\end{tabular}
\caption{Statistics of datasets. $c$: Number of authors. $D$: Number of documents. $\mu$: Median number of documents per author. $\sigma$: Standard deviation of document number. $d$: Median document size (tokens).}
\label{tab:datasets}
\end{table}

\begin{table*}[t]
\centering
\begin{tabular}{l|l||c|c|c|c|c}
\textbf{Model} & Parameter & Emails &  IMDb & Blogs & Twitter & reddit \\\hline
SCAP & ngram size & 9 & 7 & 4 & 6 & 5\\
Imposters & ngram size & 9 & 4 & 4 & 4 & 4 \\
\multirow{2}{*}{LDAH-S} & \# topics & 300 & 300 & 100 & 300 & 300 \\
                        & vocabulary size & 40,000 & 30,000 & 30,000 & 30,000 & 50,000
\end{tabular}
\caption{Hyperparameters for SCAP, Imposters, and LDAH-S for all datasets. SCAP and Imposters with profile size 14,000. Imposters with feature proportion 0.3, vocabulary size 30,000, 500 iterations. LDAH-S with 2,000 iterations.}
\label{tab:parameters}
\end{table*}

All word embedding channels are initialized with 300-dimensional GloVe vectors \cite{Pennington2014} trained on 840B tokens of the Common Crawl corpus\footnote{For the Twitter dataset 200-dimensional GloVe vectors  trained on 27B tokens of Twitter data are used.}. Character embedding channels and words not present in the set of pre-trained words are initialized randomly.

\subsection{Comparison methods}

We compare against four state-of-the-art authorship attribution methods:

\begin{itemize}
	\item \textbf{SVM+Stems}: an SVM classifier, which distinguishes authors based on word stems rather than bag-of-words \cite{Allison2008}. Features are additionally weighted with tf-idf and scaled to have unit variance.
	\item \textbf{SCAP}: the Source Code Author Profile (SCAP) method \cite{Frantzeskou2007} determines authorship based on the intersection of the most frequent character n-grams of an unknown text and an author's profile, i.e. the concatentation of an author's known texts.
	\item \textbf{Imposters}: the Imposters Method \cite{Koppel2011} is based on the intuition that the profile of an unknown text's author is likely to be most similar to the unknown text most often as the feature set varies. For each iteration, it calculates the cosine similarity of a random fraction of a feature set of space-free character n-grams between an unknown text and an author's profile. It then chooses the author who exhibited the maximum cosine similarity most frequently. 
	\item \textbf{LDAH-S}: LDA Hellinger Single-Document, the top-performing method of Seroussi et al. \shortcite{Seroussi2011} uses Hellinger distance between the LDA topic distributions for an unknown text and an author's profile as a measure for authorship.

\end{itemize}

Results for SVM+Stems have been reported on the Email dataset \cite{Allison2008} and results for Imposters and LDAH-S have been reported on the IMDb62 and Blog dataset \cite{Seroussi2011}. However, as there are no pre-defined train-test splits for these datasets, we re-implement the above algorithms to guarantee identical conditions and objective comparison.

\subsection{Hyperparameters and training}

As authorship attribution algorithms must be able to deal with a large number of authors, we conduct experiments for a large range of author numbers. For each number of authors $n$, we select the subset of texts belonging to the $n$ most authors with the most documents in the dataset. Note that as we scale the number of authors, the number of documents per author diminishes and identifying prominent authors transforms into the scenario of detecting authors with only few samples. We do not equalize the number of training documents to maintain the imbalanced distribution common in real-world applications.

As the per-fold run time of most of the state-of-the-art authorship attribution methods renders 10-fold cross-validation prohibitively expensive, we randomly split off 10\% as a stratified test set. We keep the test set constant and use a seed to reduce randomness in the comparison. We furthermore use 10\% of each training set as a stratified development set. We optimize hyperparameters for all comparison algorithms for each dataset on the development set of its 10-author subset using random search and keep them constant each dataset. For SVM+Stems, we use a linear kernel, unigram stems, and a vocabulary size of 10,000. The parameters for the other algorithms can be seen in Table \ref{tab:parameters}.

\begin{table*}[t]
\centering
\begin{tabular}{l|c|c|c|c|c|c|c|c|c||c}
\textbf{Model} & \multicolumn{2}{c|}{Emails} &  IMDb & \multicolumn{2}{c|}{Blogs} & \multicolumn{2}{c|}{Twitter} & \multicolumn{2}{c ||}{reddit} & \texttt{Ave} \\\hline
Number of authors & 10 & 50 & 62 & 10 & 50 & 10 & 50 & 10 & 50 &- \\\hline
SVM+Stems & 83.0 & 72.9 & 88.3 & 36.5 & 29.7 & 91.7 & 81.3 & 35.1 & 21.2 & 60.0  \\
SCAP & 83.1 & 69.0 & \textbf{94.8} & 48.6 & 41.6 & 91.3 & 82.5 & 46.5 & 30.3 & 65.3 \\
Imposters & 52.0 & 32.9 & 76.9 & 35.4 & 22.6 & 71.4 & 52.5 & 32.1 & 16.3 & 43.6 \\
LDAH-S & 82.0 & 39.1 & 72.0 & 52.5 & 18.3 & 90.0 & 38.3 & 43.0 & 14.2 & 49.9 \\\hline
CNN-word & 89.7 & 82.5 & 84.3 & 59.0 & 43.0 & 96.2 & 80.5 & 36.2 & 20.1 & 65.7 \\
CNN-word-word & 90.3 & 81.0 & 82.0 & 58.8 & 41.4 & 95.7 & 79.3 & 39.6 & 18.3 & 65.2 \\
CNN-char & \textbf{93.1} & \textbf{88.1} & 91.7 & 59.7 & 48.1 & \textbf{97.5} & \textbf{86.8} & \textbf{58.8} & \textbf{37.2} & \textbf{73.4} \\
CNN-word-char & 90.3 & 84.9 & 90.2 & \textbf{61.2} & \textbf{49.4} & 97.2 & 84.9 & 53.1 & 27.7 & 70.1 \\
CNN-word-word-char & 90.1 & 83.8 & 88.4 & \textbf{61.2} & 47.0 & 95.9 & 84.0 & 56.1 & 27.0 & 70.4 \\ 
\end{tabular}
\caption{Micro-averaged F1 scores of our CNN variants and state-of-the-art authorship attribution methods for 10 and 50 authors (except for IMDb) in the chosen domains. We also show a model's average performance (\texttt{Ave}) across all scores to give an estimate of how well a model generalizes across domains and consequently might perform in a new domain.}
\label{tab:results}
\end{table*}

\textbf{CNN parameters}. We optimize hyperparameters for our CNN configurations on the development set of the 10-authors emails subset without any task-specific fine-tuning. These are: vocabulary size of 10,000 words, 98 characters\footnote{We use the same vocabulary as Zhang et al. \shortcite{Zhang2015d}, but distinguish between lower-case and upper-case characters and convolve all numbers into one as both measures increase performance.}, maximum sequence length of 500 for word-based CNNs and of 3000 for CNN-char and hybrid-channel CNNs, $l_2$ constraint of 0, dropout rate of 0.5, filter windows of 6, 7, 8 with 100 feature maps each, and mini-batch size of 50. We train for 15 epochs using mini-batch stochastic gradient descent and early stopping. We use the Adadelta update rule \cite{Zeiler2012} as it allows us to pay special attention to infrequent features that can be, however, highly indicative of certain authors.

\section{Results and discussion}

\begin{figure*}[]
\centering
\begin{minipage}{.49\textwidth}
  \centering
  \includegraphics[width=\linewidth]{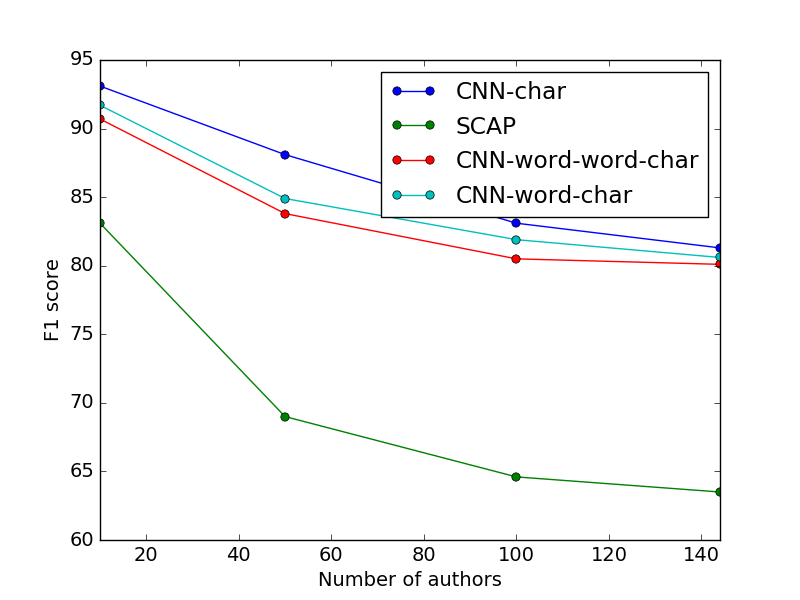}
  \captionof{figure}{F1 scores on Emails dataset}
  \label{fig:emails}
\end{minipage}
\begin{minipage}{.49\textwidth}
  \centering
  \includegraphics[width=\linewidth]{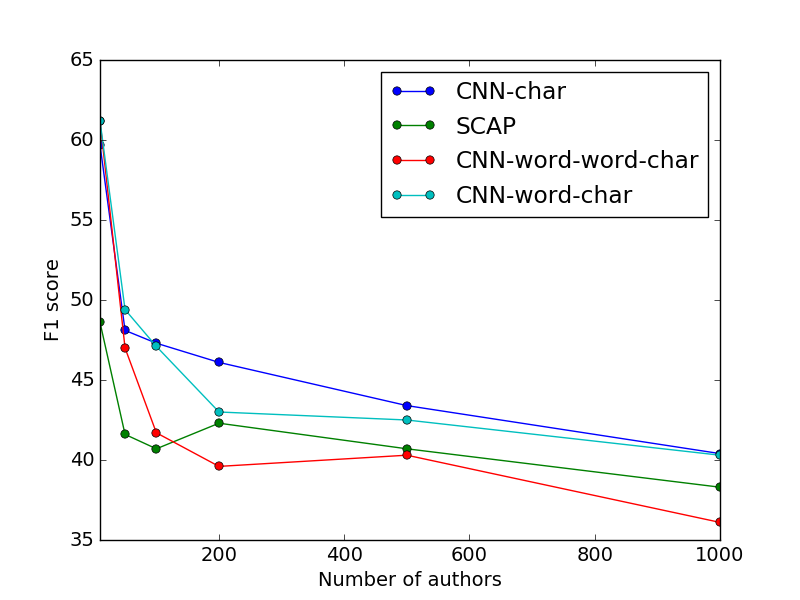}
  \captionof{figure}{F1 scores on Blogs dataset}
  \label{fig:blogs}
\end{minipage}
\end{figure*}

We evaluate all algorithms on all 62 authors for the IMDb dataset and on 10 and 50 authors respectively for all other datasets. Results of our CNN models against the comparison methods are listed in Table \ref{tab:results}. Additionally, we evaluate the best CNNs against the best comparison method in Table \ref{tab:results}, SCAP on larger numbers of authors. We show results up to all 144 authors for the Emails dataset in Figure \ref{fig:emails} and results up to 1,000 authors for the Blogs, Twitter, and reddit datasets in Figures \ref{fig:blogs}, \ref{fig:twitter}, and \ref{fig:reddit} respectively.

\subsection{Domain}

The corpus domain has a big impact on a model's performance and often requires domain-specific feature engineering \cite{Stamatatos2009}. Word-based methods such as SVM+Stems, LDAH-S, CNN-word, CNN-word-word perform well in domains in which topical information is discriminatory, such as emails, movie reviews, and blogs. They, however, achieve comparatively worse performance for short-message domains such as Twitter and reddit. On these, character-enhanced methods provide a considerable performance boost. Past studies \cite{Stamatatos2009,Schwartz2013} highlighted challenges for short text authorship attribution, such as the number of authors, the training set size, and the size of the test document.

\textbf{Twitter}. The generally impressive scores in the Twitter domain reveal a bias that is two-fold: a) As we have not pre-processed the data (other than lower-casing) to reflect real-world use cases, models are able to leverage mentions and hashtags to achieve high F1 scores; replacing these lowers scores for all models as observed by Layton \shortcite{Layton2010}; b) celebrities and power-users that are the subject of this dataset use language differently in comparison to regular users, with the goal of enhancing their brand by frequently tweeting similar key messages. This behavior, however, makes it easier for models to pick up on their individual style or preferred content. We intend to investigate the stylistic differences between regular users and celebrities and their impact on authorship attribution in future work.

\textbf{Reddit.} Most models perform poorly on the reddit domain. We suspect that authorship attribution on reddit is challenging for two reasons: a) Comments are either very long (tirades / explanations) or very short (short replies / puns); b) in contrast to Twitter, reddit is less about broadcasting oneself to the world and more about interacting with the community; posts thus often reflect the character of the thread rather than the character of the user and often contain stylistically conspicuous features such as irony \cite{Wallace2015}. Further research may thus reveal how users shift their style and if this behavior differs across different subreddits.

\subsection{CNNs}

All of our CNN variants consistently outperform most traditional authorship attribution methods. Even CNN-word that only uses word embeddings performs significantly better than most comparison methods. Moreover, CNNs show aptitude to handle the class imbalance problem \cite{Stamatatos2007} inherent in real-world applications by significantly outperforming the comparison methods on all datasets with imbalanced numbers of documents per author (see Table \ref{tab:datasets}) in line with findings by Potthast et al. \shortcite{Potthast}.

In the emails domain, CNNs outperform the comparison methods by more than 6 \%. for 10 authors and SCAP by more than 16 \%. for 144 authors. We suppose that these large performance differences lie in the fact that CNNs are able to pay special attention to structural measures such as greetings, farewells, and signatures \cite{DeVel2001}. Differences for the IMDb domain are less pronounced, as authors would generally review similar movies, rendering specific words or character sequences discriminatory. Fine-tuned word embeddings that are sensitive to topical divergence between authors boost CNN performance in the blogs domain. They are, however, less helpful in the Twitter and reddit domains, where hashtags or emoticons are the most characteristic features.

\begin{figure*}[]
\centering
\begin{minipage}{.49\textwidth}
  \centering
  \includegraphics[width=\linewidth]{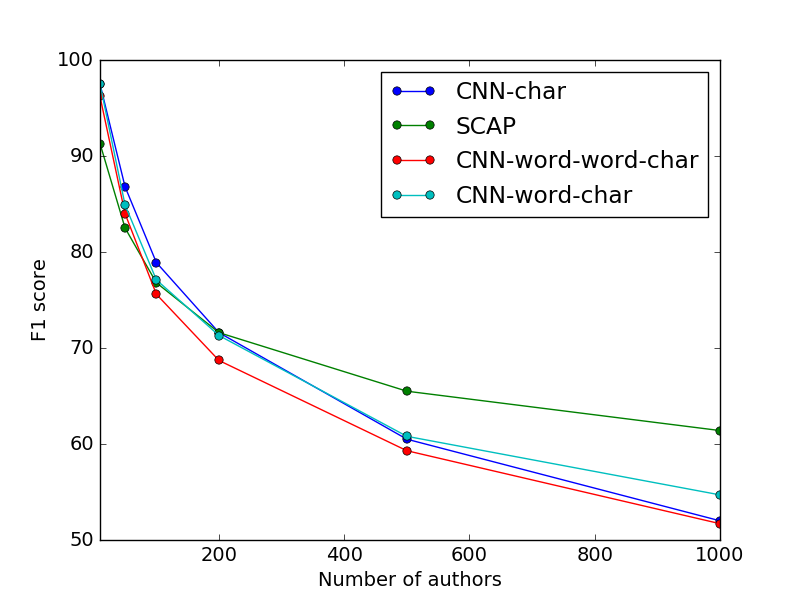}
  \captionof{figure}{F1 scores on Twitter dataset}
  \label{fig:twitter}
\end{minipage}
\begin{minipage}{.49\textwidth}
  \centering
  \includegraphics[width=\linewidth]{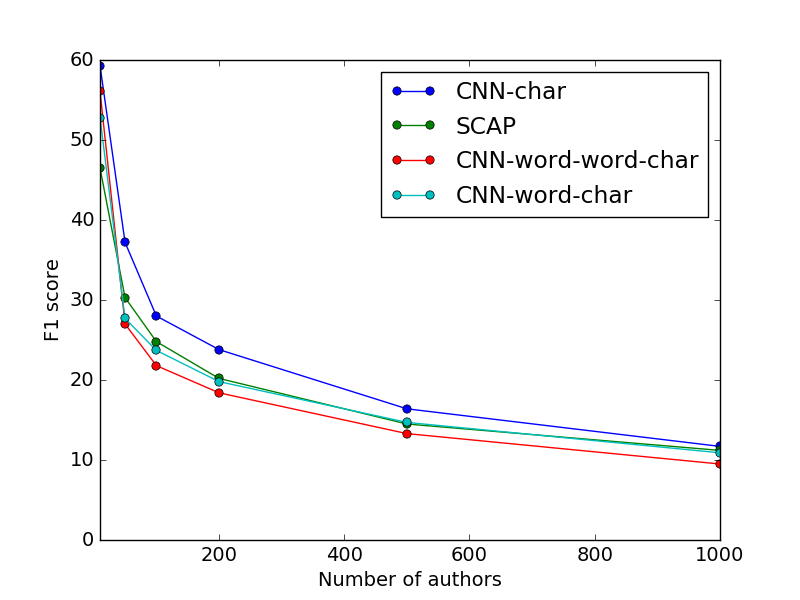}
  \captionof{figure}{F1 scores on Reddit dataset}
  \label{fig:reddit}
\end{minipage}
\end{figure*}

\textbf{Character-level CNNs} outperform traditional methods and CNN variants on 4 out of 5 datasets in Table \ref{tab:results}, on most of them quite significantly. They outperform comparison methods in domains such as Reddit where certain character sequences such as smileys and emoticons are discriminatory by more than 12 \%. for 10 authors. They perform well even with fewer per-class examples scaling better than all other methods (except for SCAP on Twitter in Figure \ref{fig:twitter}) to large numbers of authors.

The reason for the success of character-level CNNs clearly is their superior ability to capture stylistic information encoded on the character-level. In contrast to traditional approaches such as SCAP and Imposters, character-level CNNs are able to model more complex interactions between different stylistic features by leveraging non-linearities. 

Given the availability of a GPU, they only take a few hours to train -- even for large-scale authorship attribution scenarios as the author number only affects the number of parameters in the final softmax layer, which scales linearly. More importantly, they are able to form a prediction instantaneously compared to the CPU-hours or CPU-days required by SCAP or the Imposters method respectively. For this reason, they are particularly suited for conducting a large or recurring number of predictions in on-line scenarios such as attributing messages to known terrorists \cite{Abbasi2005}.

\textbf{Hybrid-channel CNNs} outperform CNNs that rely solely on word embeddings, as character-level information is important for the task of authorship attribution. In the blogs domain, where topical information is a distinguishing feature, they outperform character-level CNNs in Table \ref{tab:results}. Using a static and a non-static word channel together with the character channel had similar effects as using a multi-channel architecture over a regular word-based CNN, i.e. it increased scores on some datasets.

\subsection{Traditional methods}

Similarly to Frantzeskou and Stamatatos \shortcite{Frantzeskou2007}, we find that increasing the profile size of an author's concatenated known texts consistently increases performance for n-gram based similarity methods, i.e. SCAP and Imposters. The optimal profile size for our datasets, 14,000, however, is considerably higher than past values reported for this hyperparameter \cite{Layton2010} \cite{Layton2012}, suggesting that larger ranges should be considered in future research.

We have found that restricting the vocabulary size by selecting only the 30,000 most frequent space-free character n-grams for the Imposters method generally increased performance as frequency is the most important criterion for selecting features in authorship attribution \cite{Stamatatos2009}.\footnote{Koppel et al. \shortcite{Koppel2011} use more than 250,000 unique character n-grams.} Even though we improve performance for Imposters on the IMDb dataset in comparison to Seroussi et al. \shortcite{Seroussi2011} by selecting appropriate hyperparameters, we are unable to achieve competitive scores using the more recently proposed authorship attribution methods, Imposters and LDAH-S.\footnote{We  apply them to larger numbers of authors and obtain similar results. Note that we evaluate -- in contrast to Koppel et al. \shortcite{Koppel2011} and Seroussi et al. \shortcite{Seroussi2011} -- using F1, which penalizes low recall.} In contrast, the SCAP method proves to be easy and fast to train and outperforms CNNs on the IMDb dataset in Table \ref{tab:results}.

Impressively, SCAP scales better than CNNs on Twitter in Figure \ref{fig:twitter}. Recall that for Twitter influencers, certain n-grams such as user mentions or hashtags are highly indicative. We suspect that as SCAP stores discrete n-grams for each author, it is better able to assign them to the correct author, while the continuous function used by CNNs might blur the boundaries in the case of many authors. To mitigate this deficit, CNNs can a) be made more expressive by adding more layers \cite{Zhang2015d}; or b) can be interpolated with n-grams that capture clear n-gram-author correspondences as in the Twitter domain.

\section{Conclusion}

In this work, we have applied character-level CNNs to large-scale authorship attribution. We have extensively evaluated combinations of different CNN input channels and introduced a novel model that combines character and word channels to leverage both stylistic and topical information. We have compared CNNs against state-of-the-art methods for a large range of author numbers, shedding new light on traditional approaches. We have presented state-of-the-art results for four out of five datasets in different domains and have introduced two new Twitter and reddit datasets that we make available for further research.

\section*{Acknowledgments}

This publication has emanated from research supported by Grant Number EBPPG/2014/30 from the Irish Research Council with Aylien Ltd. as Enterprise Partner and by Grant Number SFI/12/RC/2289 from Science Foundation Ireland (SFI).

\bibliography{acl2016_authorship}
\bibliographystyle{acl2016}

\end{document}